\documentclass[runningheads]{llncs}
\usepackage{graphicx}
\usepackage{amsmath}
\usepackage{amssymb}
\usepackage{booktabs}
\usepackage{relsize}
\usepackage[table]{xcolor}

\usepackage{hyperref}

\usepackage{verbatim}

\begin{document}
\title{Quantifying the Intrinsic Usefulness of Attributional Explanations for Graph Neural Networks with Artificial Simulatability Studies}
\titlerunning{Quantifying Explanation Usefulness for Graph Neural Networks}

\newif\ifanon

\ifanon
\author{Anonymous}
\authorrunning{Anon et al.}
\institute{
Anonymous Institute\\
\email{\{abc,lncs\}@anon.edu}}

\else
\author{
Jonas Teufel\inst{1}\orcidID{0000-0002-9228-9395} \and
Luca Torresi\inst{1}\orcidID{0000-0003-2205-6753} \and
Pascal Friederich\thanks{corresponding author}\inst{1}\orcidID{0000-0003-4465-1465}}
\authorrunning{
J. Teufel et al.
}
\institute{
Institute of Theoretical Informatics (ITI), Karlsruhe Institute of Technology (KIT), Karlsruhe, Germany\\
\email{\{jonas.teufel\}@student.kit.edu}\\
\email{\{luca.torresi,pascal.friederich\}@kit.edu}}
\fi


\maketitle              
\begin{abstract}

Despite the increasing relevance of explainable AI, assessing the quality of explanations remains a challenging issue. Due to the high costs associated with human-subject experiments, various proxy metrics are often used to approximately quantify explanation quality. Generally, one possible interpretation of the quality of an explanation is its inherent value for teaching a related concept to a student. In this work, we extend artificial simulatability studies to the domain of graph neural networks. Instead of costly human trials, we use explanation-supervisable graph neural networks to perform simulatability studies to quantify the inherent \textit{usefulness} of attributional graph explanations. We perform an extensive ablation study to investigate the conditions under which the proposed analyses are most meaningful. We additionally validate our method's applicability on real-world graph classification and regression datasets. We find that relevant explanations can significantly boost the sample efficiency of graph neural networks and analyze the robustness towards noise and bias in the explanations. We believe that the notion of usefulness obtained from our proposed simulatability analysis provides a dimension of explanation quality that is largely orthogonal to the common practice of faithfulness and has great potential to expand the toolbox of explanation quality assessments, specifically for graph explanations.

\end{abstract}
%
%
%


\section{Introduction}

Explainable AI (XAI) methods are meant to provide explanations alongside a complex model's predictions to make its inner workings more transparent to human operators to improve trust and reliability, provide tools for retrospective model analysis, as well as to comply with anti-discrimination laws \cite{doshi-velez_towards_2017}. Despite recent developments and a growing corpus of XAI methods, a recurring challenge remains the question of how to assess the quality of the generated explanations.

\noindent Since explainability methods aim to improve human understanding of complex models, Doshi-Velez and Kim \cite{doshi-velez_towards_2017} argue that ultimately the quality of explanations has to be assessed in a human context. To accomplish this, the authors propose the idea of simulatability studies. In that context, human subjects are tasked to simulate the behavior of a machine-learning model given different amounts of information. While a control group of participants receives only the model input-output information, the test group additionally receives the explanations in question. If, in that case, the test group performs significantly better at simulating the behavior, the explanations can be assumed to contain information useful to human understanding of the task. However, human trials such as this are costly and time-consuming, especially considering the number of participants required to obtain a statistically significant result. Therefore, the majority of XAI research is centered around more easily available proxy metrics such as explanation sparsity and faithfulness.

\noindent While proxy metrics are an integral part of the XAI evaluation pipeline, we argue that the quantification of usefulness obtained through simulatability studies is an important next step toward comparing XAI methods and thus increasing the impact of explainable AI. Recently, Pruthi \textit{et al.} \cite{pruthi_evaluating_2021} introduce the concept of \textit{artificial simulatability studies} as a trade-off between cost and meaningfulness. Instead of using human subjects, the authors use explanation-supervisable neural networks as participants to conduct simulatability studies for natural language processing tasks.\\

\noindent In this work, we extend the concept of artificial simulatability studies to the domain of graph neural networks and specifically node and edge attributional explanations thereof. This application has only been enabled through the recent development of sufficiently explanation-supervisable graph neural network approaches \cite{teufel_megan_2022}. We will henceforth refer to this artificial simulatability approach as the student-teacher analysis of explanation quality: The explanations in question are considered to be the "teachers" that are evaluated on their effectiveness of communicating additional task-related information to explanation-supervisable "student" models. We show that, under the right circumstances, explanation supervision leads to significantly improved main task prediction performance w.r.t. to a reference. We first conduct an extensive ablation study on a specifically designed synthetic dataset to highlight the conditions under which this effect can be optimally observed. Most importantly, we find that the underlying student model architecture has to be sufficiently capable to learn explanations during explanation-supervised training. Our experiments show, that this is especially the case for the self-explaining MEGAN architecture, which was recently introduced by Teufel \textit{et al.} \cite{teufel_megan_2022}.

\noindent Additionally, we find that the target prediction problem needs to be sufficiently challenging to the student models to see a significant effect. We can furthermore show that while ground truth explanations cause an increase in performance, deterministically incorrect/adversarial explanations cause a significant decrease in performance. In the same context, random explanation noise merely diminishes the benefit of explanations, but neither causes a significant advantage nor a disadvantage.

\noindent Finally, we validate the applicability of our method on explanations for one real-world molecular classification and one molecular regression dataset.
\nopagebreak
\nopagebreak
\section{Related Work}



\subsubsection{Simulatability Studies.}


Doshi-Velez and Kim \cite{doshi-velez_towards_2017} introduce the concept of simulatability studies, in which human participants are asked to simulate the forward predictive behavior of a given model. Explanations about the model behavior should be considered useful if a group of participants with access to these explanations performs significantly better than a control group without them. Such studies are only rarely found in the growing corpus of XAI literature due to the high effort and cost associated with them. Nonetheless, some examples of such studies can be found. Chandrasekaran \textit{et al.} \cite{chandrasekaran_explanations_2018} for example conduct a simulatability study for a visual question answering (VQA) task. The authors investigate the effect of several different XAI methods such as GradCAM and attention among other aspects. They find no significant performance difference for participants when providing explanations. Hase and Bansal \cite{hase_evaluating_2020} conduct a simulatability study for a sentiment classification task. They can only report significant improvements for a small subset of explanation methods. Lai \textit{et al.} \cite{lai_human_2019,lai_why_2020} conduct a simulatability study for a deception detection task. Unlike previously mentioned studies, the authors ask participants to predict ground truth labels instead of simulating a model's predictions. Among different explanation methods, they also investigate the effects of other assistive methods on human performance, such as procedurally generated pre-task tutorials and real-time feedback. The study shows that real-time feedback is crucial to improve human performance. In regard to explanations, the authors find that especially simplistic explanations methods seem to be more useful than more complicated deep-learning-based ones and that providing the polarity of attributional explanations is essential.\\
Beyond the cost and effort associated with human trials, previous studies report various additional challenges when working with human subjects. One issue seems to be the limited working memory of humans, where participants report forgetting previously seen relevant examples along the way. Another issue is the heterogeneity of participants' abilities, which causes a higher variance in performance results, necessitating larger sample sizes to obtain statistically significant results. Overall, various factors contribute to such studies either not observing any effect at all or reporting only on marginal explanation benefits.\\
One possible way to address this is proposed by Arora \textit{et al.} \cite{arora_explain_2022}, who argue to rethink the concept of simulatability studies itself. In their work, instead of merely using human subjects as passive predictors, the participants are encouraged to interactively engage with the system. In addition to guessing the model prediction, participants are asked to make subsequent single edits to the input text with the goal of maximizing the difference in model confidence. The metric of the average confidence deviation per edit can then also be seen as a measure of human understanding of the model's inner workings. The authors argue that such an explorative and interactive study design is generally more suited to the strengths of human subjects and avoids their respective weaknesses.\\
Another approach is represented by the emergent idea of \textit{artificial simulatability studies}, which generally aim to substitute human participants in these kinds of studies with machine learning models that are able to learn from explanations in a similar manner. There exist early variations of this basic idea  \cite{hase_leakage-adjusted_2020,treviso_explanation_2020}, for which conceptional problems have been pointed out \cite{pruthi_evaluating_2021}. Most notably, some methods expose explanations during test time, which may cause label leakage. Recently, Pruthi \textit{et al.} \cite{pruthi_evaluating_2021} devise a method that does not expose explanations during test time by leveraging explanation-supervised model training. They are able to show a statistically significant test performance benefit for various explanation methods, as well as for explanations derived from human experts in natural language processing tasks.
In our work, we build on the basic methodology proposed by Pruthi \textit{et al.} and use explanation-supervisable student models to avoid the label-leakage problem. Furthermore, we extend their basic approach toward a more rigorous method. The authors consider the \textit{absolute} performance of the explanation supervised student by itself as an indicator of simulatability. We argue that, due to the stochastic nature of neural network training, potential simulatability benefits should only be considered on a statistical level obtained through multiple independent repetitions, only \textit{relative} to a direct reference, and verified by tests of statistical significance.

\subsubsection{Explanation Supervision for GNNs} 

Artificial simulatability studies, as previously discussed, require student models which are capable of \textit{explanation supervision}. This means that it should be possible to directly train the generated explanations to match some given ground truth explanations during the model training phase. Explanation supervision has already been successfully applied in the domains of image processing \cite{linsley_learning_2019} and natural language processing \cite{boyd_cyborg_2022}. However, only recently was the practice successfully adapted to the domain of graph neural networks as well. First, Gao \textit{et al.} \cite{gao_gnes_2021} propose the GNES framework, which aims to use the differentiable nature of various existing post-hoc explanation methods such as GradCAM and LRP to perform explanation supervised training. Teufel \textit{et al.} \cite{teufel_megan_2022} on the other hand introduce the MEGAN architecture which is a specialized attention-based architecture showing especially high potential for explanation-supervision. To the best of our knowledge, these two methods remain the only existing methods for explanation-supervision of graph \textit{attributional} explanations until now.\\
In addition to attributional explanations, several other types of explanations have been introduced. Noteworthy examples are prototype-based explanations \cite{shin_page_2022} and concept-based explanations \cite{magister_gcexplainer_2021}. In the realm of prototype explanations, Zhang \textit{et al.} \cite{zhang_protgnn_2022} and Dai and Wang \cite{dai_towards_2021} introduce self-explaining prototype-based graph neural networks, although it has not yet been demonstrated if and how explanation-supervision could be applied to them. For concept-based explanations, on the other hand, Magister \textit{et al.} \cite{magister_encoding_2022} demonstrate explanation supervision, opening up the possibility to extend artificial simulatability studies to explanation modalities beyond simple attributional explanations as well.

\section{Student-Teacher Analysis of Explanation Quality}


\begin{figure}[b!]
    \centering
    \includegraphics[width=\textwidth]{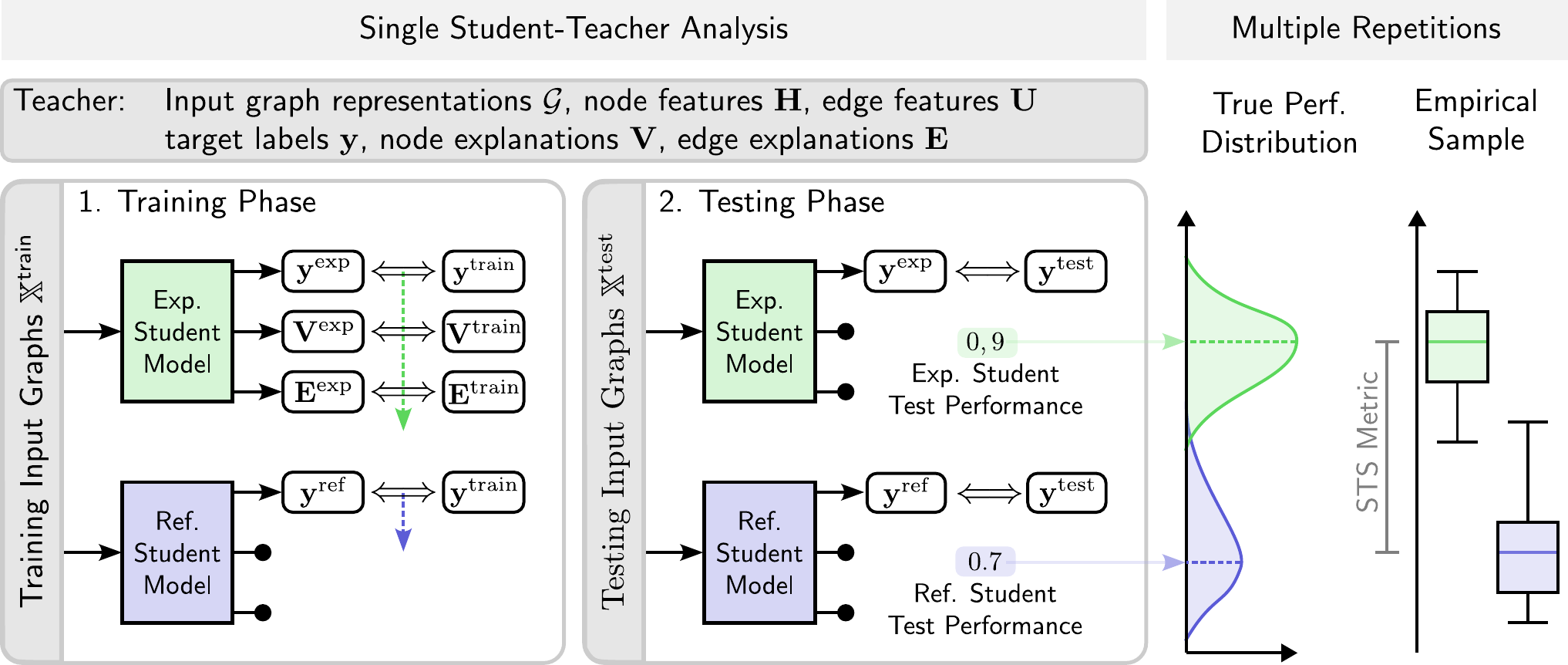}
    \caption{Illustration of the student teacher training workflow as well as the setting of our artificial simulatability study.}
    \label{fig:concept}
\end{figure}

Simulatability studies aim to assess how useful a set of explanations is in improving human understanding of a related task. To offset the high cost and uncertainty associated with human-subject experiments, Pruthi \textit{et al.} \cite{pruthi_evaluating_2021} introduce artificial simulatability studies, which substitute human participants with explanation-aware neural networks, for natural language processing tasks. In this section, we describe our extension of this principle idea to the application domain of graph neural networks and introduce the novel STS metric which we use to quantify the explanation-induced performance benefit.\\
  
\noindent We assume a directed graph $\mathcal{G} = (\mathcal{V}, \mathcal{V})$ is represented by a set of node indices $\mathcal{V} = \mathbb{N}^{V}$ and a set of edges $\mathcal{E} \subseteq \mathcal{V} \times \mathcal{V}$, where a tuple $(i, j) \in \mathcal{E}$ denotes a directed edge from node $i$ to node $j$. Every node $i$ is associated with a vector of initial node features $\mathbf{h}^{(0)}_i \in \mathbb{R}^{N_0}$, combining into the initial node feature tensor $\mathbf{H}^{(0)} \in \mathbb{R}^{V \times N_0}$. Each edge is associated with an edge feature vector $\mathbf{u}^{(0)}_{ij} \in \mathbb{R}^{M}$, combining into the edge feature tensor $\mathbf{U} \in \mathbb{R}^{E \times M}$. Each graph is also annotated with a target value vector $\mathbf{y}^{\text{true}} \in \mathbb{R}^{C}$, which is either a one-hot encoded vector for classification problems or a vector of continuous values for regression problems.  For each graph exists node and edge attributional explanations in the form of a node importance tensor $\mathbf{V} \in [0, 1]^{V \times K}$ and an edge importance tensor $\mathbf{E} \in [0, 1]^{E \times K}$ respectively. $K$ is the number of explanation channels and is usually equal to the size $C$ of the target vector, meaning that for every target value each element of the input graph is annotated with a 0 to 1 value indicating that element's importance.
\\

\noindent In the framework of artificial simulatability studies, human participants are replaced by explanation-aware machine learning models which will be referred to as \textit{students}. In this analogy, the \textit{teacher} is represented by the dataset of input graphs and target value annotations, as well as the explanations whose quality is to be determined. Figure~\ref{fig:concept} illustrates the concept of such a \textit{student-teacher analysis} of explanation quality. The set $\mathbb{X}$ of input data consists of tuples $(G, \textbf{H}^{(0)}, \textbf{U}^{(0)})$ of graphs and their features. The set $\mathbb{Y}$ consists of tuples $(\mathbf{y}, \mathbf{V}, \mathbf{E})$ of target value annotations, as well as node and edge attributional explanations. A student is defined as a parametric model $\mathcal{S}_{\theta}: (G, \textbf{H}^{(0)}, \textbf{U}^{(0)}) \rightarrow (\mathbf{y}, \mathbf{V}, \mathbf{E})$ with the trainable model parameters ${\pmb \theta}$. This firstly implies that every student model has to directly output explanations alongside each prediction. Moreover, these generated explanations have to be actively \textit{supervisable} to qualify as an explanation-aware student model.\\
During a single iteration of the student-teacher analysis, the sets of input and corresponding output data are split into a training set $\mathbb{X}^{\text{train}}, \mathbb{Y}^{\text{train}}$ and an unseen test set $\mathbb{X}^{\text{test}}, \mathbb{Y}^{\text{test}}$ respectively. Furthermore, two architecturally identical student models are initialized with the same initial model parameters ${\pmb\theta}$: The reference student model $\mathcal{S}^{\text{ref}}_{\theta}$ and the explanation-aware student model $\mathcal{S}^{\text{exp}}_{\theta}$. During the subsequent training phase, the reference student only gets to train on the main target value annotations $\mathbf{y}$, while the explanation student is additionally trained on the given explanations. After the two students were trained on the same training elements and the same hyperparameters, their final prediction performance is evaluated on the unseen test data. If the explanation student outperforms the reference student on the final evaluation, we can assume that the given explanations contain additional task-related information and can thus be considered useful in this context.\\
However, the training of complex models, such as neural networks, is a stochastic process that generally only converges to a local optimum. For this reason, a single execution of the previously described process is not sufficient to assess a possible performance difference. Rather, a repeated execution is required to confirm the statistical significance of any result. Therefore, we define the student-teacher analysis as the $R$ repetitions of the previously described process, resulting in the two vectors of test set evaluation performances $\textbf{p}^{\text{ref}}, \textbf{p}^{\text{exp}} \in \mathbb{R}^{R}$ for the two student models respectively. The concrete type of metric used to determine the final performance may differ, as is the case with classification and regression problems for example. Based on this definition we define the \textit{student-teacher simulatability} metric
\begin{equation*}
\operatorname{STS}_{R} = \text{median}( \textbf{p}^{\text{exp}} - \textbf{p}^{\text{ref}} )
\end{equation*}
as the median of the pairwise performance differences between all the individual explanation students' and reference students' evaluation results. We choose the median here instead of the arithmetic mean, due to its robustness towards outliers, which may occur when models sporadically fail to properly converge in certain iterations of the procedure.\\
In addition to the calculation of the $\operatorname{STS}$ metric, a paired t-test is performed to assure the statistical significance of the results. Only if the p-value of this test is below a 5\% significance level should the analysis results be considered meaningful.




\section{Computational Experiments}

\subsection{Ablation Study for a Synthetic Graph Classification Dataset}

\begin{figure}[b!]
    \centering
    \includegraphics[width=\textwidth]{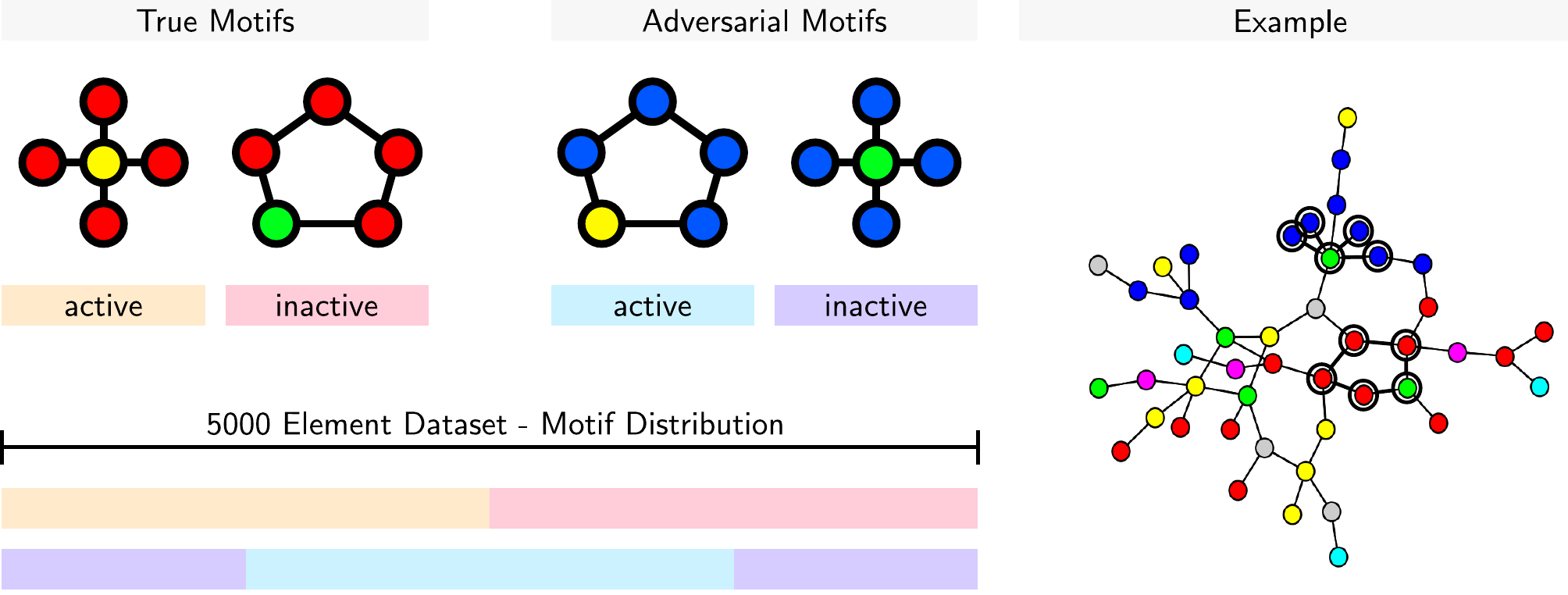}
    \caption{Synthetic dataset used to quantify the usefulness of attributional graph explanations, incl. testing the robustness toward adversarial explanations.}
    \label{fig:dataset}
\end{figure}

We first conduct an ablation study on a specifically designed synthetic graph dataset to show the special conditions under which a performance benefit for the explanation student can be observed.\\
We call the synthetic dataset created for this purpose \textit{red and blue adversarial motifs} and a visualization of it can be seen in Figure~\ref{fig:dataset}. The dataset consists of 5000 randomly generated graphs where each node is associated with 3 node features representing an RGB color code. Each graph is seeded with one primarily red motif: Half of the elements are seeded with the red and yellow star motif and are consequently labeled as the "active" class. The other half of the elements are seeded with a red and green ring motif and labeled as "inactive". The dataset represents a binary classification problem where each graph will have to be classified as either active or inactive. As each class assignment is entirely based on the existence of the corresponding sub-graph motifs, these motifs are considered the perfect ground truth explanations for that dataset. In addition to the primarily red motifs, each graph is also seeded with one primarily blue motif: Either a blue-yellow ring motif or a blue-green star motif. These blue motifs are seeded such that their distribution is completely uncorrelated with the true class label of the elements. Thus, these motifs are considered deterministically incorrect/adversarial explanations w.r.t. the main classification task.
%
%
\subsubsection{Student Model Implementations. }

We conduct an experiment to assess the suitability of different student model implementations. As previously explained, a student model has to possess two main properties: Node and edge explanations have to be generated alongside each prediction and more importantly it has to be possible to train the models based on these explanations in a supervised manner. To the best of our knowledge, there exist two methods from literature, which do this for \textit{attributional} explanations: The GNES framework of Gao \textit{et al.} \cite{gao_gnes_2021} and the MEGAN architecture of Teufel \textit{et al.} \cite{teufel_megan_2022}. We conduct an experiment with $R = 25$ repetitions of the student-teacher analysis for three different models: A lightweight MEGAN model, GNES explanations based on a simple GCN network, and GNES explanations based on a simple GATv2 network. In each iteration, 100 elements of the dataset are used to train the student model while the rest is used during testing. Table~\ref{tab:gnes_results} shows the results of this experiment. We report the final $\operatorname{STS}$ value, as well as the node and edge AUC metrics, which indicate how well the explanations of the corresponding models match the ground truth explanations of the test set.\\

\begin{table}
    \caption{Results for 25 repetitions of the student-teacher analysis for different reference models (Ref) and explanation supervised student model (Exp) implementations.}
    \label{tab:gnes_results}
    \renewcommand*{\arraystretch}{1.4}
\setlength\tabcolsep{8pt}
\newcolumntype{g}{>{\columncolor{green!3}}c}
\newcolumntype{b}{>{\columncolor{blue!3}}c}
\begin{center}
\begin{tabular}{lcbgbg}
\toprule
\multicolumn{1}{c}{Student Model} &
\multicolumn{1}{c}{$\operatorname{STS}_{25} \uparrow$} &
\multicolumn{2}{c}{$\text{Node AUC} \uparrow$} &
\multicolumn{2}{c}{$\text{Edge AUC} \uparrow$} \\
&
&
Ref &
Exp &
Ref &
Exp \\
\midrule

$\text{GNES}_{\text{GCN}}$ &
$0.02$ &
$0.55 { \color{gray} \mathsmaller{ \pm 0.04 } }$ &
$0.59 { \color{gray} \mathsmaller{ \pm 0.03 } }$ &
$0.64 { \color{gray} \mathsmaller{ \pm 0.04 } }$ &
$0.66 { \color{gray} \mathsmaller{ \pm 0.04 } }$ \\

$\text{GNES}_{\text{GATv2}}$ &
$0.01$ &
$0.59 { \color{gray} \mathsmaller{ \pm 0.05 } }$ &
$0.61 { \color{gray} \mathsmaller{ \pm 0.05 } }$ &
$0.51 { \color{gray} \mathsmaller{ \pm 0.05 } }$ &
$0.55 { \color{gray} \mathsmaller{ \pm 0.04 } }$ \\

$\text{MEGAN}^{2}_{0.0}$ &
$\mathbf{0.12}^{(*)}$ &
$0.64 { \color{gray} \mathsmaller{ \pm 0.15 } }$ &
$\mathbf{0.94} { \color{gray} \mathsmaller{ \pm 0.01 } }$ &
$0.66 { \color{gray} \mathsmaller{ \pm 0.14 } }$ &
$\mathbf{0.96} { \color{gray} \mathsmaller{ \pm 0.02 } }$ \\

\bottomrule

\end{tabular}
\end{center}
    \smaller \hspace*{2pt} $^{\text{(*)}}$ Statistically significant according to a paired T-test with $p<5\%$
\end{table}

\noindent Since the perfect ground truth explanations are used for this experiment, we expect the explanation student to have the maximum possible advantage w.r.t to the explanations. The results show that only the MEGAN student indicates a statistically significant $\operatorname{STS}$ value of a median 12\% accuracy improvement for the explanation-aware student. The GNES experiments on the other hand do not show statistically significant performance benefits. We believe that this is due to the limited effect of the explanation supervision that can be observed in these cases: While the node and edge accuracy of the GNES explanation student only improves by a few percent, the MEGAN explanation student almost perfectly learns the ground truth explanations. This is consistent with the results reported by Teufel \textit{et al.} \cite{teufel_megan_2022}, who report that MEGAN outperforms the GNES approach in capability for explanation supervision. A possible explanation for why that is the case might be that the explanation-supervised training of the already gradient-based explanations of GNES relies on a second derivative of the network, which might provide a generally weaker influence on the network's weights.\\
Based on this result, we only investigate the MEGAN student in subsequent experiments.


\begin{figure}[t]
    \centering
    \includegraphics[width=\textwidth]{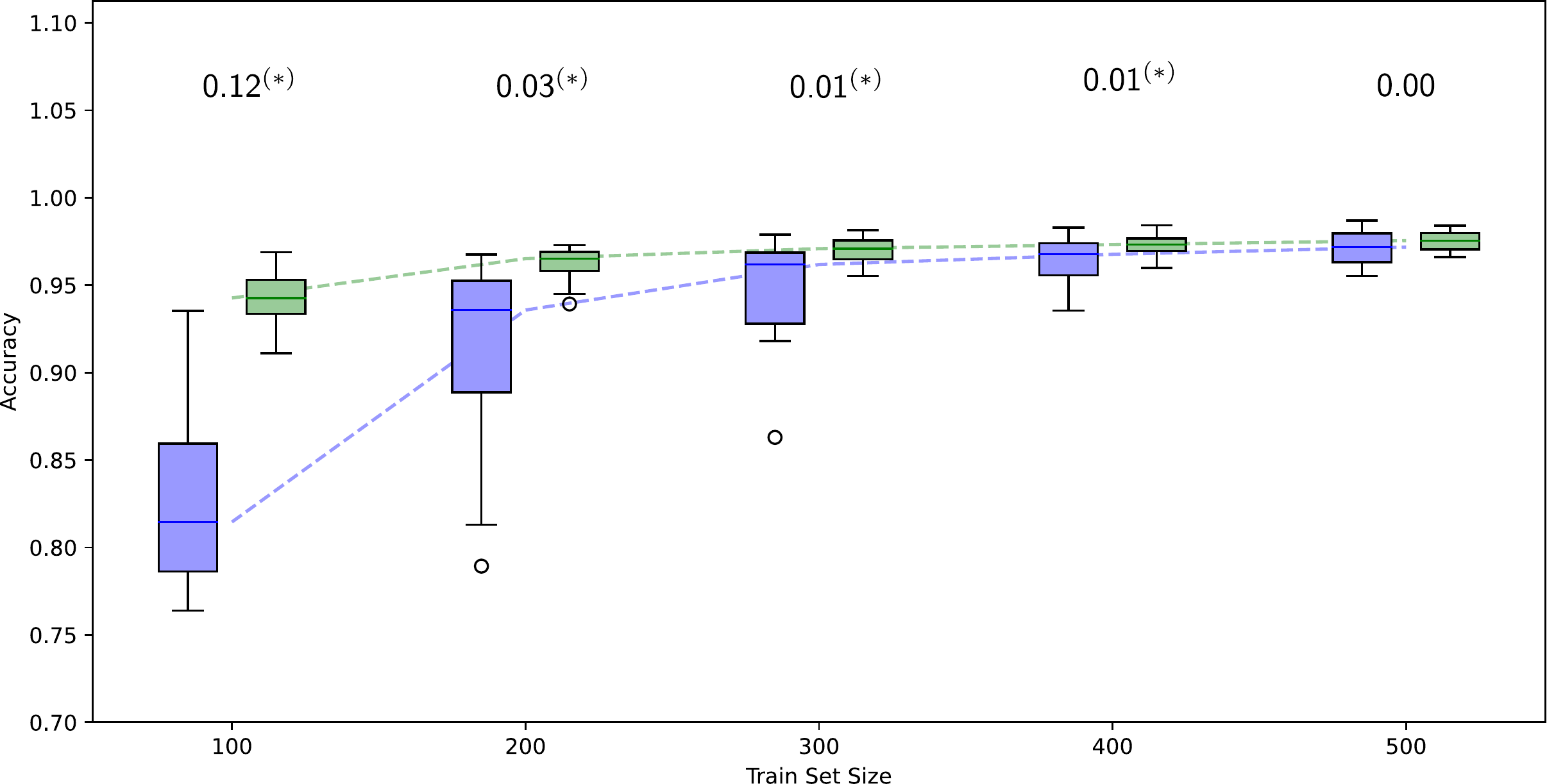}
    \caption{Results of student-teacher analyses ($R=25$) for different training dataset sizes. Each column shows the performance distribution for the reference student ({\color{blue} blue}) and the explanation student ({\color{olive} green}) of the student-teacher procedure. The number above each column is the resulting $\operatorname{STS}$ value. $\text{(*)}$ indicates statistical significance according to a paired T-test with $p<5\%$}
    \label{fig:train_size_results}
\end{figure}

\begin{figure}[t]
    \centering
    \includegraphics[width=\textwidth]{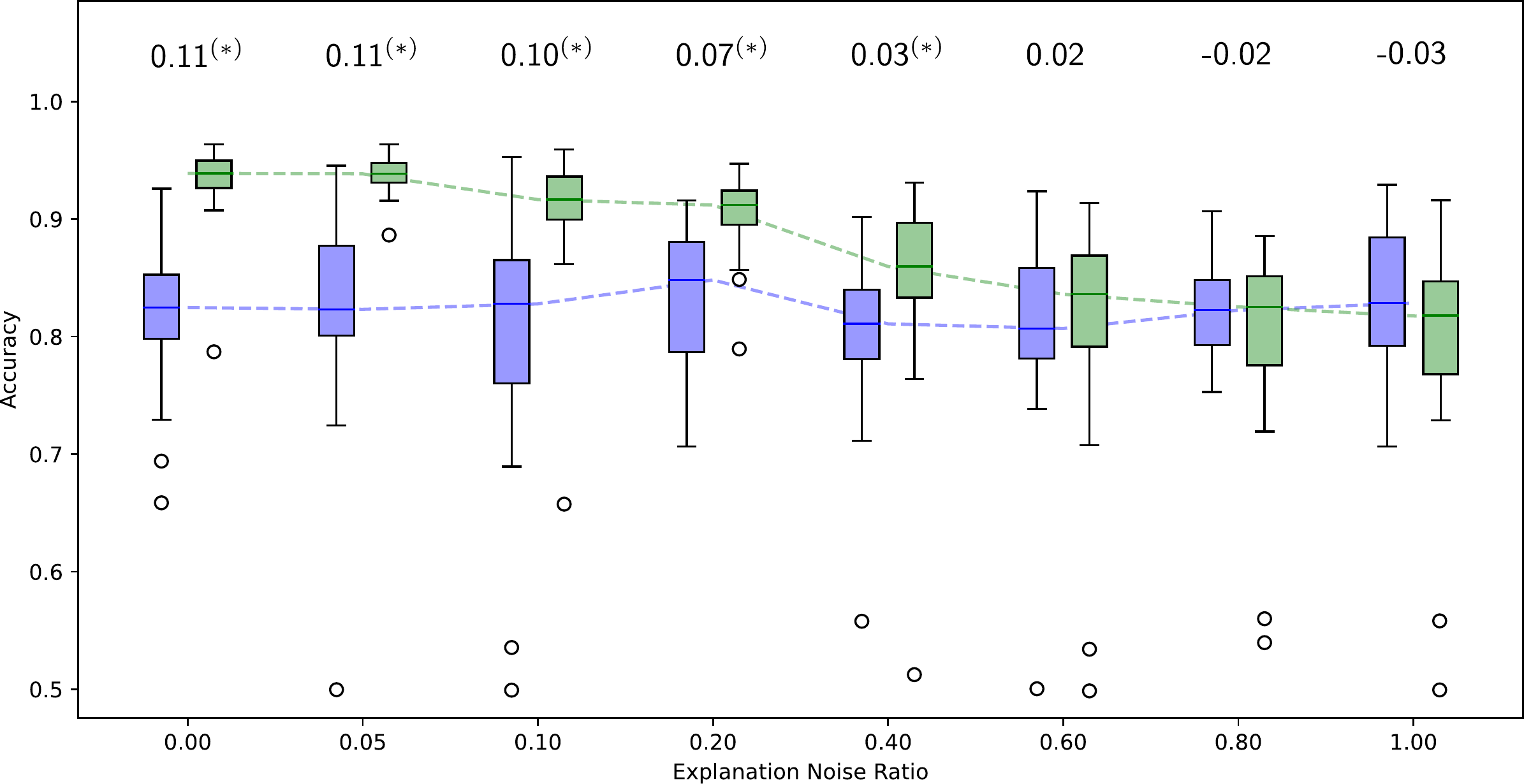}
    \caption{Results of student-teacher analyses ($R=25$) for explanations with different ratios of additional explanation noise. Each column shows the performance distribution for the reference student ({\color{blue} blue}) and the explanation student ({\color{olive} green}) of the student-teacher procedure. The number above each column is the resulting $\operatorname{STS}$ value. $\text{(*)}$ indicates statistical significance according to a paired T-test with $p<5\%$}
    \label{fig:noise_results}
\end{figure}

\begin{figure}[t]
    \centering
    \includegraphics[width=\textwidth]{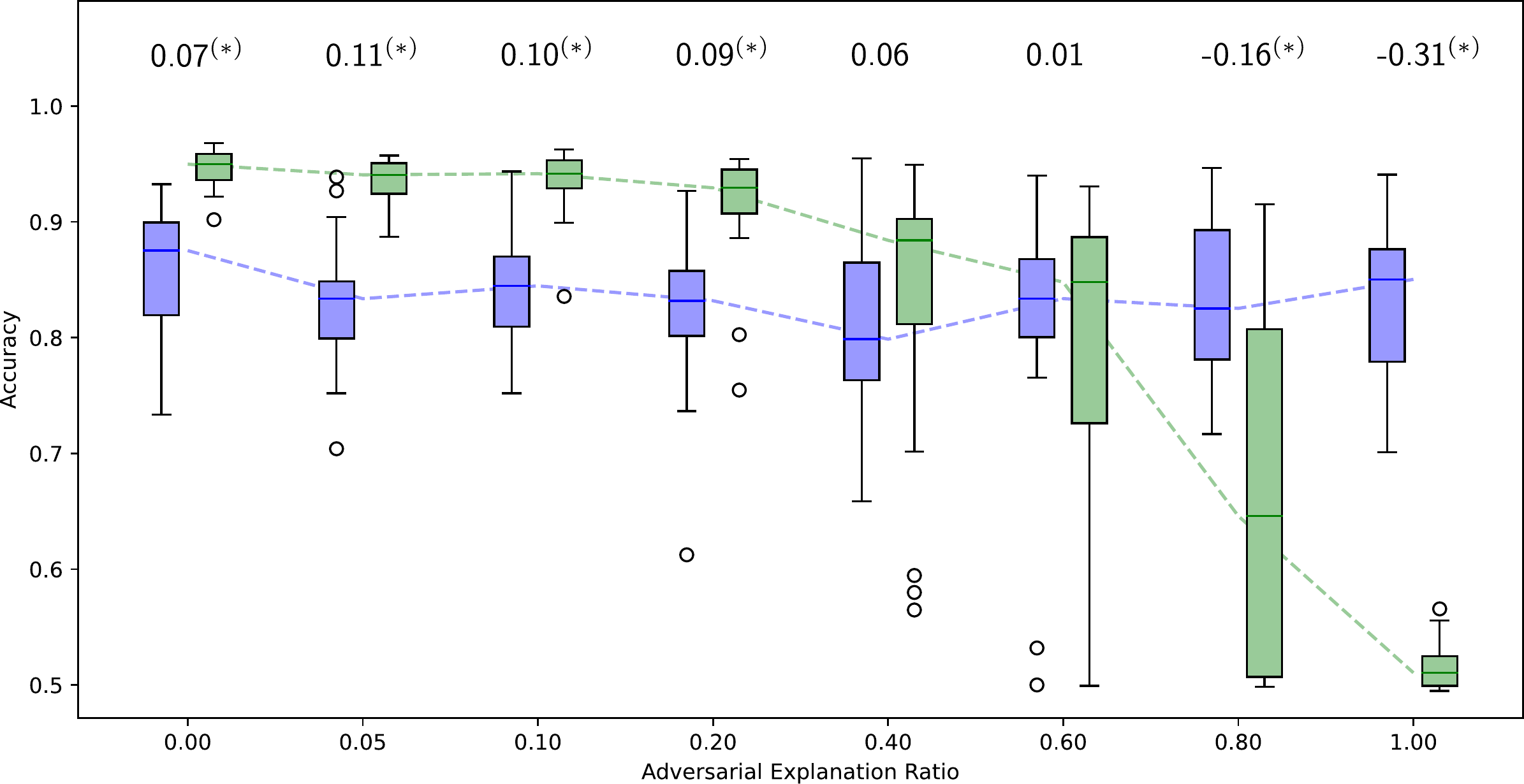}
    \caption{Results of student-teacher analyses ($R=25$) for datasets containing different amounts of adversarial incorrect explanations. Each column shows the performance distribution for the reference student ({\color{blue} blue}) and the explanation student ({\color{olive} green}) of the student-teacher procedure. The number above each column is the resulting $\operatorname{STS}$ value. $\text{(*)}$ indicates statistical significance according to a paired T-test with $p<5\%$}
    \label{fig:adversarial_results}
\end{figure}

\begin{figure}[t]
    \centering
    \includegraphics[width=\textwidth]{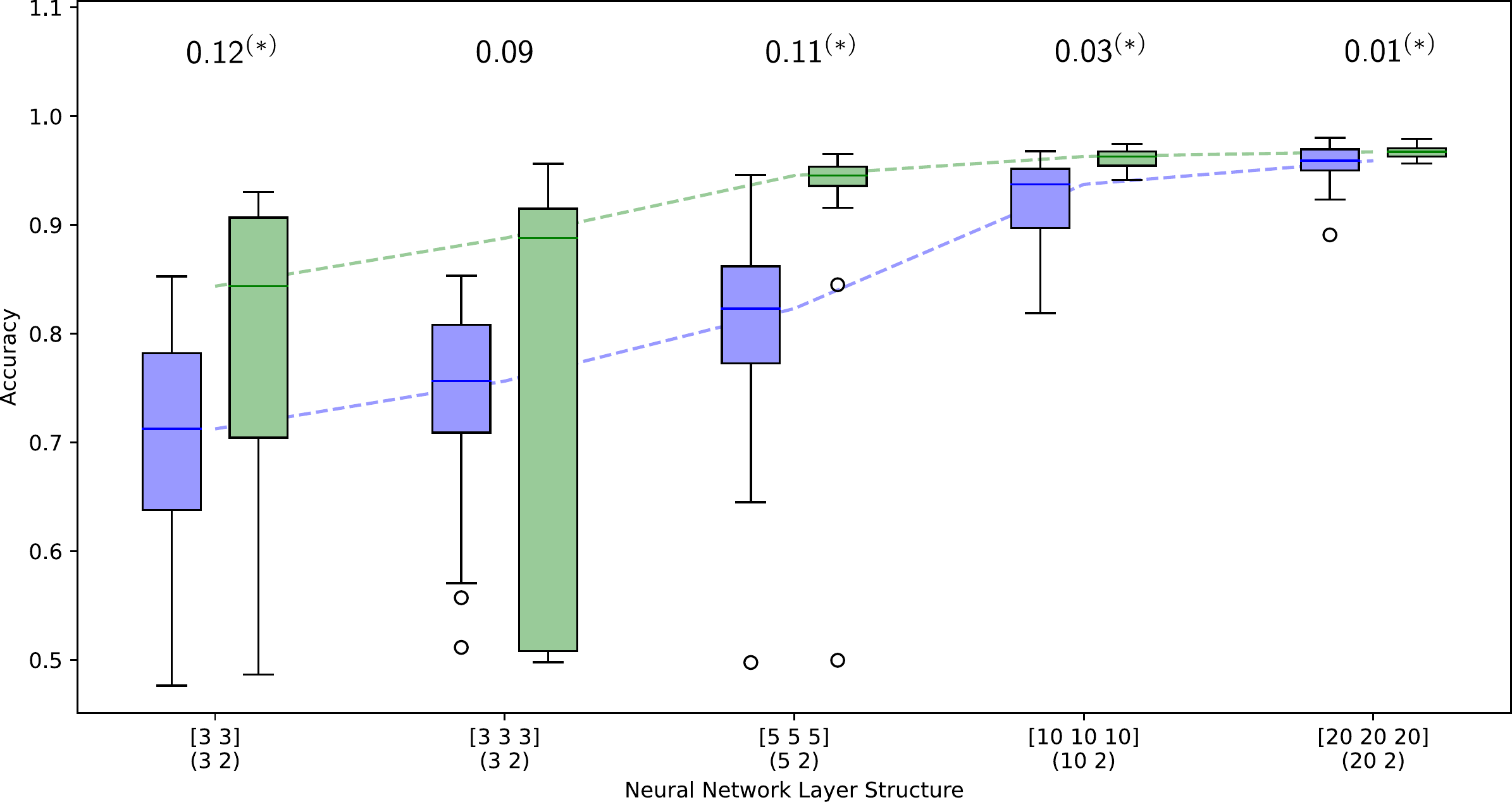}
    \caption{Results of student-teacher analyses ($R=25$) for different layer structures of the MEGAN student model. The square brackets indicate the number of hidden units in each layer of the main convolutional part of the network. The normal brackets beneath indicate the number of hidden units in the fully connected layers in the tail-end of the network. Each column shows the performance distribution for the reference student ({\color{blue} blue}) and the explanation student ({\color{olive} green}) of the student-teacher procedure. The number above each column is the resulting $\operatorname{STS}$ value. $\text{(*)}$ indicates statistical significance according to a paired T-test with $p<5\%$}
    \label{fig:layer_results}
\end{figure}


\subsubsection{Training Dataset Size Sweep.}

In this experiment, we investigate the influence of the training dataset size on the explanation performance benefit. For this purpose, we conduct several student-teacher analyses with $R = 25$ repetitions using the MEGAN student architecture. We vary the number of elements used for training between 100, 200, 300, 400, and 500 elements out of a total of 5000. In each iteration, the training dataset with that number of elements is randomly sampled from the entire dataset and the rest is used during testing. Figure~\ref{fig:train_size_results} shows the results of this experiment. We visualize the performance distributions of explanation and reference students for each dataset size and provide the $\operatorname{STS}$ metric in each case.\\

\noindent The results show the greatest performance benefit for the smallest training set size of just 100 elements. Afterward, the $\operatorname{STS}$ value converges to 0 for 500 elements, losing statistical significance as well. We believe that this is caused by the convergence of \textit{both} students to the near-perfect performance of approx. 98\% accuracy. In other words: A larger train set size represents a smaller difficulty for the student models. With decreasing difficulty, the students can solve the task almost perfectly by themselves, diminishing any possible benefit of the explanations. We can therefore formulate the rule of thumb that explanations have the potential to provide the greatest benefit when tasks are \textit{more difficult}, and cannot be so easily solved without explanations. As shown in this experiment, a reduction of the train set size sufficiently provides such an increase in difficulty.\\
Based on this result, we conduct subsequent experiments with a training set size of 100 to observe the most pronounced effect.


\subsubsection{Explanation Noise Sweep.}

For the majority of real-world tasks, perfect ground truth explanations are generally not available. Instead, explanations can be generated through a multitude of XAI methods that have been proposed in recent years. Since complex machine learning models and XAI methods generally only find local optima, it is reasonable to assume that generated explanations are not perfect but rather contain some amount of noise as well. The question is how such explanation noise affects the results of our proposed student-teacher analysis. In this experiment, we perform different student-teacher analyses, where in each case the explanations are overlaid with a certain ratio $P\%$ of random noise, where $P \in \{0, 5, 10, 20, 40, 60, 80, 100\}$. A ratio $P\%$ means that the explanation importance value for every element (nodes and edges) in every graph has a $P\%$ chance of being randomly sampled instead of the ground truth value being used. Each student-teacher analysis is conducted with a MEGAN student architecture and 100 training data points.
Figure~\ref{fig:train_size_results} shows the results of this experiment.\\

\noindent The results show that there is a statistically significant performance benefit for the explanation student until 40\% explanation noise is reached. Afterward, the $\operatorname{STS}$ value converges towards zero and loses statistical significance as well. One important aspect to note is that even for high ratios of explanation noise the performance difference converges toward zero. This indicates that explanations consisting almost entirely of \textit{random noise} do not benefit the performance of a student model, but they do \textit{not negatively influence} it either. We believe this is the case because random explanations do not cause any learning effect for the model. In our setup of explanation-supervised training, actual explanation labels are not accessible to either student during the testing phase, instead, the models have to learn to replicate the given explanations during training through their own internal explanation-generating mechanisms. Only through these learned replications can any potential advantage or disadvantage be experienced by the models during performance evaluation. Completely random explanations cannot be learned by the models and consequently have no effect during performance evaluation.


\subsubsection{Adversarial Explanation Sweep.}

The previous experiment indicates that purely random explanations do not negatively affect the model performance. By contrast, it could be expected that deterministic incorrect explanations on the other hand should have a negative influence on the performance. The used dataset is seeded with two families of sub-graph motifs (see Figure~\ref{fig:dataset}): The red-based motifs are completely correlated with the two target classes and can thus be considered the perfect explanations for the classification task. The blue-based motifs on the other hand are completely uncorrelated to the task and can thus be considered \textit{incorrect/adversarial} explanations w.r.t. to the target labels. In this experiment, increasing amounts of these adversarial explanations are used to substitute the true explanations during the student-teacher analysis to investigate the effect of incorrect explanations on the performance difference. In each iteration,  $Q\%$ of the true explanations are replaced by adversarial explanations, where $Q \in \{0, 5, 10, 20, 40, 60, 80, 100\}$. Each student-teacher analysis is conducted with a MEGAN student architecture and 100 training elements.\\

\noindent The results in Figure~\ref{fig:adversarial_results} show that a statistically significant explanation performance benefit remains for ratios of adversarial explanations for up to 20\%. For increasingly large ratios, the $\operatorname{STS}$ value still remains positive although the statistical significance is lost. For ratios of 80\% and above, statistically significant \textit{negative} $\operatorname{STS}$ values can be observed. This implies that incorrect explanations negatively influence the performance of the explanation-aware student model. 



\subsubsection{Student Network Layer Structure.}
In this experiment, we investigate the influence of the concrete student network layout on the explanation performance benefit. For this purpose, we conduct several student-teacher analyses with $R = 25$ repetitions using the MEGAN student architecture. We vary the number of convolutional and fully-connected layers, as well as the number of hidden units in these layers. Starting with a simple two-layer 3-unit network layout, the number of model parameters, and thus its complexity is gradually increased until the most complex case of a three-layer 20-unit network is reached. Figure~\ref{fig:train_size_results} shows the results of this experiment. We visualize the performance distributions of explanation and reference students for each dataset size and provide the $\operatorname{STS}$ metric in each case.\\

\noindent The results show that the students' prediction performance generally improves for more complex models. However, this is true for the explanation as well as the reference student. While there still is a statistically significant effect for the most complex network layout, it is very marginal because the reference student achieves almost perfect accuracy in these cases as well. On the other hand, the most simple student network layout shows the largest performance benefit. However, for the simple network layouts, the standard variation of the performance over the various repetitions is greatly increased for reference and explanation students, but seemingly more so for the explanation student. We generally conclude that both extreme cases of simplistic and complex student network architectures have disadvantages w.r.t. to revealing a possible explanation performance benefit. In the end, the best choice is a trade-off between variance in performance and overall capability. 


\subsubsection{Node versus Edge Explanations.}

We conduct an experiment to determine the relative impact of the node and edge explanations individually. We conduct a student-teacher analysis with $R=25$ repetitions. We use a simple three-layer MEGAN student, where each iteration uses 100 randomly chosen training samples. We investigate three cases: As a baseline case, the explanation student uses ground truth node and edge explanations during explanation-supervised training. In another case, the explanation student is only supplied with the node attributional explanations. In the last case, only the edge attributional explanations are used. This is achieved by setting the corresponding weighting factors to 0 during training. Table~\ref{tab:node_edge_results} shows the results of this experiment. We report the final $\operatorname{STS}$ value, as well as the node and edge AUC metrics, which indicate how well the explanations of the corresponding models match the ground truth explanations of the test set.\\

\begin{table}
    \caption{Results for 25 repetitions of the student-teacher Analysis conducted with either only node explanations, only edge explanations, or both.}
    \label{tab:node_edge_results}
    \renewcommand*{\arraystretch}{1.4}
\setlength\tabcolsep{8pt}
\newcolumntype{g}{>{\columncolor{green!3}}c}
\newcolumntype{b}{>{\columncolor{blue!3}}c}
\begin{center}
\begin{tabular}{lcbgbg}
\toprule
\multicolumn{1}{c}{Explanations} &
\multicolumn{1}{c}{$STS_{25} \uparrow$} &
\multicolumn{2}{c}{$\text{Node AUC} \uparrow$} &
\multicolumn{2}{c}{$\text{Edge AUC} \uparrow$} \\
&
&
Ref &
Exp &
Ref &
Exp \\
\midrule

$\text{Both}$ &
$0.12^{(*)}$ &
$0.62 { \color{gray} \mathsmaller{ \pm 0.14 } }$ &
$0.95 { \color{gray} \mathsmaller{ \pm 0.03 } }$ &
$0.62 { \color{gray} \mathsmaller{ \pm 0.16 } }$ &
$0.94 { \color{gray} \mathsmaller{ \pm 0.03 } }$ \\

$\text{Nodes}$ &
$0.12^{(*)}$ &
$0.65 { \color{gray} \mathsmaller{ \pm 0.13 } }$ &
$0.93 { \color{gray} \mathsmaller{ \pm 0.03 } }$ &
$0.65 { \color{gray} \mathsmaller{ \pm 0.12 } }$ &
$0.92 { \color{gray} \mathsmaller{ \pm 0.04 } }$ \\

$\text{Edges}$ &
$0.10^{(*)}$ &
$0.67 { \color{gray} \mathsmaller{ \pm 0.15 } }$ &
$0.93 { \color{gray} \mathsmaller{ \pm 0.03 } }$ &
$0.67 { \color{gray} \mathsmaller{ \pm 0.12 } }$ &
$0.94 { \color{gray} \mathsmaller{ \pm 0.03 } }$ \\

\bottomrule

\end{tabular}
\end{center}
    \smaller \hspace*{2pt} $^{\text{(*)}}$ Statistically significant according to a paired T-test with $p<5\%$
\end{table}

\noindent The results show that all three cases achieve statistically significant $\operatorname{STS}$ values indicating a performance benefit of the given explanations. Furthermore, in all three cases, the explanations learned by the explanation student show high similarity (AUC $> 0.9$) to the ground truth explanations for node \textit{as well as} edge attributions. This implies that the student model is able to infer the corresponding explanation edges for the ground truth explanatory motifs, even if it is only trained on the nodes, and vice versa. We believe the extent of this property is a consequence of the used MEGAN student architecture. The MEGAN network architecture implements an explicit architectural co-dependency of node and edge explanations to promote the creation of connected explanatory sub-graphs. These results imply that it may be possible to also apply the student-teacher analysis in situations where only node or edge explanations are available.


\subsection{Real-World Datasets}

In addition to the experiments on the synthetic dataset, we aim to provide a validation of the student-teacher analysis' effectiveness on real-world datasets as well. For this purpose, we choose one graph classification and one graph regression dataset from the application domain of chemistry. We show how the student-teacher analysis can be used to quantify \textit{usefulness} of the various kinds of explanations for these datasets.

\subsubsection{Mutagenicity - Graph Classification}

To demonstrate the student-teacher analysis of GNN-generated explanations on a real-world graph classification task, we choose the Mutagenicity dataset \cite{hansen_benchmark_2009} as the starting point. By its nature of being real-world data, this dataset does not have ground truth explanations as it is, making it hard to compare GNN-generated explanations to the ground truth. However, the dataset can be transformed into a dataset with ground truth explanatory subgraph motifs. It is hypothesized that the nitro group ($\text{NO}_2$) is one of the main reasons for the property of mutagenicity \cite{lin_generative_2021,luo_parameterized_2020}. Following the procedure previously proposed by Tan \textit{et al.} \cite{tan_learning_2022}, we extract a subset of elements containing all molecules which are labeled as mutagenic and contain the benzene-$\text{NO}_2$ group as well as all the elements that are labeled as non-mutagenic and do not contain that group. Consequently, for the resulting mutagenicity subset, the benzene-$\text{NO}_2$ group can be considered as the definitive ground truth explanation for the mutagenic class label. We call the resulting dataset \textit{MutagenicityExp}. It consists of roughly 3500 molecular graphs, where about 700 are labeled as mutagenic. Furthermore, we designate 500 random elements as the test set, which are sampled to achieve a balanced label distribution.\\

\noindent Based on this dataset, we train GNN models to solve the classification problem. Additionally, we use multiple different XAI methods to generate attributional explanations for the predictions of those GNNs on the previously mentioned test set of 500 elements. These explanations, generated by the various XAI methods, are then subjected to student-teacher analysis, along with some baseline explanations. The results of an analysis with 25 repetitions can be found in Table~\ref{tab:mutagenicity_results}.\\
The hyperparameters of the student-teacher analysis have been chosen through a brief manual search. We use the same basic three-layer MEGAN student architecture as with the synthetic experiments. In each repetition, 10 random elements are used to train the students, and the remainder is used to assess the final test performance. Each training process employs a batch size of 10, 150 epochs, and a 0.01 learning rate. The student-teacher analysis is performed solely on the previously mentioned 500-element test set, which remained unseen to any of the trained GNN models. 

\begin{table}
    \caption{Results for 25 repetitions of the student-teacher analysis for different explanations on the MutagenicityExp dataset. We mark the best result in bold and underline the second best.}
\label{tab:mutagenicity_results}
\renewcommand*{\arraystretch}{1.4}
\setlength\tabcolsep{8pt}
\newcolumntype{g}{>{\columncolor{green!3}}c}
\newcolumntype{b}{>{\columncolor{blue!3}}c}
\begin{center}
\begin{tabular}{lcbgbg}
\toprule
\multicolumn{1}{c}{Explanations by} &
\multicolumn{1}{c}{$\operatorname{STS}_{25} \uparrow$} &
\multicolumn{2}{c}{$\text{Node AUC} \uparrow$} &
\multicolumn{2}{c}{$\text{Edge AUC} \uparrow$} \\
&
&
Ref &
Exp &
Ref &
Exp \\
\midrule

$\text{Ground Truth}$ &
$\mathbf{0.13}^{(*)}$ &
$0.42 { \color{gray} \mathsmaller{ \pm 0.05 } }$ &
$\mathbf{0.97} { \color{gray} \mathsmaller{ \pm 0.05 } }$ &
$0.41 { \color{gray} \mathsmaller{ \pm 0.05 } }$ &
$\mathbf{0.96} { \color{gray} \mathsmaller{ \pm 0.04 } }$ \\

$\text{GNNExplainer}$ &
$0.09^{(*)}$ &
$0.50 { \color{gray} \mathsmaller{ \pm 0.09 } }$ &
$0.69 { \color{gray} \mathsmaller{ \pm 0.05 } }$ &
$0.50 { \color{gray} \mathsmaller{ \pm 0.11 } }$ &
$0.71 { \color{gray} \mathsmaller{ \pm 0.04 } }$ \\

$\text{Gradient}$ &
$0.07^{(*)}$ &
$0.54 { \color{gray} \mathsmaller{ \pm 0.18 } }$ &
$0.84 { \color{gray} \mathsmaller{ \pm 0.06 } }$ &
$0.46 { \color{gray} \mathsmaller{ \pm 0.17 } }$ &
$0.67 { \color{gray} \mathsmaller{ \pm 0.10 } }$ \\

$\text{MEGAN}^{2}_{1.0}$ &
$\underline{0.12}^{(*)}$ &
$0.55 { \color{gray} \mathsmaller{ \pm 0.15 } }$ &
$\underline{0.91} { \color{gray} \mathsmaller{ \pm 0.01 } }$ &
$0.55 { \color{gray} \mathsmaller{ \pm 0.14 } }$ &
$\underline{0.92} { \color{gray} \mathsmaller{ \pm 0.02 } }$ \\

$\text{Random}$ &
$0.01$ &
$0.50 { \color{gray} \mathsmaller{ \pm 0.04 } }$ &
$0.50 { \color{gray} \mathsmaller{ \pm 0.03 } }$ &
$0.50 { \color{gray} \mathsmaller{ \pm 0.04 } }$ &
$0.50 { \color{gray} \mathsmaller{ \pm 0.04 } }$ \\

\bottomrule

\end{tabular}
\end{center}
    \smaller \hspace*{2pt} $^{\text{(*)}}$ Statistically significant according to a paired T-test with $p<5\%$
\end{table}

\noindent As expected, the results show that the reference random explanations do not produce a statistically significant $\operatorname{STS}$ result. These explanations are included as a baseline sanity check because previous experiments on the synthetic dataset imply that purely random explanation noise should not have any statistically significant effect on the performance in either direction. The benzene-$\text{NO}_2$ ground truth explanations on the other hand show the largest statistically significant $\operatorname{STS}$ value of a median 13\% accuracy improvement, as well as the largest explanation accuracy of the explanation student models. GNNexplainer and Gradient explanations also show statistically significant STS values of 9\% and 7\% median accuracy improvement respectively. The MEGAN-generated explanations show the overall second-best results with an STS value just slightly below the ground truth.\\
We hypothesize that high values of explanation accuracy are a necessary but not sufficient condition for high STS results. A higher learned explanation accuracy indicates that the explanations are generally based on a more consistent set of underlying rules and can consequently be replicated more easily by the student network, which is the basic prerequisite to show any kind of effect during the student evaluation phase. This is a necessary but not sufficient condition because as shown in the previous adversarial explanation experiment, explanations can be highly deterministic yet conceptionally incorrect and thus harmful to model performance.

\subsubsection{AqSolDB - Graph Regression}

The AqSolDB \cite{sorkun_aqsoldb_2019} dataset consists of roughly 10000 molecular graphs annotated with experimentally determined logS values for their corresponding solubility in water. Of these, we designate 1000 random elements as the test set.\\
For the concept of water solubility, there exist no definitive attributional explanations. However, there exists some approximate intuition as to what molecular structures should result in higher/lower solubility values: In a simplified manner, one can say that non-polar substructures such as carbon rings and long carbon chains generally result in lower solubility values, while polar structures such as certain nitrogen and oxygen groups are associated with higher solubility values.\\

\noindent Based on this dataset, we train a large MEGAN model on the training split of the elements to regress the water solubility and then generate the dual-channel attributional explanations for the previously mentioned 1000-element test split. For this experiment, we only use a MEGAN model as it is the only XAI method able to create dual-channel explanations for single value graph regression tasks \cite{teufel_megan_2022}. These dual-channel explanations take the previously mentioned \textit{polarity of evidence} into account, where some substructures have an opposing influence on the solubility value. The first explanation channel contains all negatively influencing sub-graph motifs, while the second channel contains the positively influencing motifs. In addition to the MEGAN-generated explanations, we provide two baseline explanation types. Random explanations consist of randomly generated binary node and edge masks with the same shape. Trivial explanations represent the most simple implementation of the previously introduced human intuition about water solubility: The first channel contains all carbon atoms as explanations and the second channel contains all oxygen and nitrogen atoms as explanations.\\
The hyperparameters of the student-teacher analysis have been chosen through a brief manual search. We use the same basic three-layer MEGAN student architecture as with the synthetic experiments. In each repetition, 300 random elements are used to train the students, and the remainder is used to assess the final test performance. Each training process employs a batch size of 32, 150 epochs, and a 0.01 learning rate. The student-teacher analysis is performed solely on the previously mentioned 1000-element test set, which remained unseen to the predictive model during training.\\

\begin{table}
    \caption{Results for 25 repetitions of the student-teacher analysis for different explanations on the AqSolDB dataset. We highlight the best result in bold and underline the second best.}
\label{tab:sqsoldb_results}
\renewcommand*{\arraystretch}{1.4}
\setlength\tabcolsep{8pt}
\newcolumntype{g}{>{\columncolor{green!3}}c}
\newcolumntype{b}{>{\columncolor{blue!3}}c}
\begin{center}
\begin{tabular}{lcbgbg}
\toprule
\multicolumn{1}{c}{Model} &
\multicolumn{1}{c}{$STS_{25} \uparrow$} &
\multicolumn{2}{c}{$\text{Node AUC} \uparrow$} &
\multicolumn{2}{c}{$\text{Edge AUC} \uparrow$} \\
&
&
Ref &
Exp &
Ref &
Exp \\
\midrule

$\text{Random}$ &
$0.00$ &
$0.50 { \color{gray} \mathsmaller{ \pm 0.04 } }$ &
$0.50 { \color{gray} \mathsmaller{ \pm 0.03 } }$ &
$0.50 { \color{gray} \mathsmaller{ \pm 0.04 } }$ &
$0.50 { \color{gray} \mathsmaller{ \pm 0.04 } }$ \\

$\text{Trivial}$ &
$\underline{0.03}$ &
$0.40 { \color{gray} \mathsmaller{ \pm 0.05 } }$ &
$\mathbf{0.99} { \color{gray} \mathsmaller{ \pm 0.05 } }$ &
$0.42 { \color{gray} \mathsmaller{ \pm 0.05 } }$ &
$\mathbf{0.99} { \color{gray} \mathsmaller{ \pm 0.04 } }$ \\

$\text{MEGAN}^{2}_{1.0}$ &
$\mathbf{0.23}^{(*)}$ &
$0.55 { \color{gray} \mathsmaller{ \pm 0.15 } }$ &
$\underline{0.90} { \color{gray} \mathsmaller{ \pm 0.01 } }$ &
$0.55 { \color{gray} \mathsmaller{ \pm 0.14 } }$ &
$\underline{0.89} { \color{gray} \mathsmaller{ \pm 0.02 } }$ \\

\bottomrule

\end{tabular}
\end{center}
    \smaller \hspace*{2pt} $^{\text{(*)}}$ Statistically significant according to a paired T-test with $p<5\%$
\end{table}

\noindent The results show that neither the random nor the trivial explanations result in any significant performance improvement. The MEGAN-generated explanations on the other hand result in a significant improvement of a median 0.23 for the final prediction MSE. This implies that the MEGAN-generated explanations do in fact encode additional task-related information, which goes beyond the most trivial intuition about the task. However, a possible pitfall w.r.t. to this conclusion needs to be pointed out: The MEGAN-generated explanations are evaluated by a MEGAN-based student architecture. It could be that the effect is so strong because these explanations are especially well suited to that kind of architecture, as they were generated through the same architecture. We believe that previous experiments involving architecture-independent ground truth explanations have weakened this argument to an extent. Still, it will be prudent to compare these results with explanations of a different origin in the future, such as the explanations of human experts.

\section{Limitations}

We propose the student-teacher analysis as a means to measure the content of \textit{useful} task-related information contained within a set of attributional graph explanations. This methodology is inspired by human simulatability studies but with the decisive advantages of being vastly more time- and cost-efficient as well as being more reproducible. However, there are currently also some limitations to the applicability of this approach. Firstly, the approach is currently limited to attributional explanations, which assign a 0 to 1 importance value to each element. These kinds of explanations have been found to have issues \cite{kindermans_reliability_2019,adebayo_sanity_2018} and recently many different kinds of explanations have been proposed. Some examples are \textit{counterfactuals} \cite{prado-romero_gretel_2022}, \textit{concept-based} explanations \cite{magister_gcexplainer_2021}, and \textit{prototype-based} explanations \cite{shin_page_2022}.\\
Another limitation is that the student-teacher analysis process itself depends on a lot of parameters. As we show in previous sections, the size of the training dataset and the specific student architectures have an impact on how pronounced the effect can be observed. For these reasons, the proposed $\operatorname{STS}$ metric cannot be used as an absolute measure of quality such as accuracy for example. Rather, it can be used to relatively \textit{compare} different sets of explanations under the condition that all experiments are conducted with the same parameters. We propose certain rules of thumb for the selection of these parameters, however, it may still be necessary to conduct a cursory parameter search for each specific application.\\ 
Despite these limitations, we believe that artificial simulatability studies, as proposed in this work, are still an important step toward better practices for the evaluation of explainable AI methods. The currently most widespread metric of explanation quality is the concept of explanation \textit{faithfulness}, which only measures how decisive an explanation is for a model's prediction. We argue, that the concept of artificial simulatability is a first step towards a measure of how intrinsically \textit{useful} explanation can be for the \textit{communication} of additional task-related information.


\section{Conclusion}

In this work, we extend the concept of artificial simulatability studies to the application domain of graph classification and regression tasks. We propose the student-teacher analysis and the \textit{student-teacher simulatability} ($\operatorname{STS}$) metric to quantify the content of intrinsically \textit{useful} task-related information for a given set of node and edge attributional explanations. We conduct an ablation study on a synthetic dataset to investigate the conditions under which an explanation benefit can be observed most clearly and propose several rules of thumb for an initial choice of experimental parameters: Analysis requires a sufficient number of repetitions for statistical significance, a small number of training elements and a light-weight layer structure for the student model. Furthermore, we show evidence that the analysis method is robust towards small amounts of explanation noise and adversarial explanations. Interestingly, random explanation noise merely suppresses any explanation benefit while deterministically incorrect explanations cause significant performance degradation. This indicates that the method cannot only be used to identify good explanations but also to detect actively harmful ones. Furthermore, we can validate the applicability of our proposed analysis for several real-world datasets of molecular classification and regression.\\
We believe that artificial simulatability studies can provide a valuable additional tool for the evaluation of graph explanations. The student-teacher analysis measures the \textit{usefulness} of explanations in communicating task-related knowledge, which can be seen as a complementary dimension to the current widespread practice of measuring explanation faithfulness.\\

\noindent For future work, it will be interesting to extend this process to other kinds of graph explanations that have recently emerged such as concept-based explanations or prototype-based explanations. Since this is a method of measuring the content of task-related information within explanations, another application may be in educational science. The method could be used to assess explanation annotations created by human students to provide quantitative feedback on their understanding of a given graph-related problem. Another line of future work is demonstrated by Fernandes \textit{et al.} \cite{fernandes_learning_2022} which uses the differentiable nature of Pruthi \textit{et al.}'s \cite{pruthi_evaluating_2021} original artificial simulatability procedure itself in a meta-optimization process that attempts to optimize an explanation generator for this property of explanation usefulness.

\section{Reproducibility Statement}

We make our experimental code publically available at \url{https://github.com/aimat-lab/gnn_student_teacher}. The code is implemented in the Python 3.9 programming language. Our neural networks are built with the KGCNN library by Reiser \textit{et al.} \cite{reiser_graph_2021}, which provides a framework for graph neural network implementations with TensorFlow and Keras. We make all data used in our experiments publically available on a file share provider \url{https://bwsyncandshare.kit.edu/s/E3MynrfQsLAHzJC}. The datasets can be loaded, processed, and visualized with the visual graph datasets package \url{https://github.com/aimat-lab/visual_graph_datasets}. All experiments were performed on a system with the following specifications: Ubuntu 22.04 operating system, Ryzen 9 5900 processor, RTX 2060 graphics card and 80GB of memory. We have aimed to package the various experiments as independent modules and our code repository contains a brief explanation of how these can be executed.

%
%
%

\clearpage
\bibliographystyle{splncs04}
\noindent \bibliography{student_teacher}

\end{document}